\documentclass[11pt,a4paper]{article}
\usepackage[hyperref]{emnlp2018}
\usepackage{times}
\usepackage{latexsym}
\usepackage{booktabs}
\usepackage{multirow}
\usepackage{colortbl}
\usepackage{makecell}
\usepackage{enumitem}
\usepackage{array}

\makeatletter
\newcommand\showfsize[1]{{#1 size: \f@size / \f@baselineskip}}
\makeatother

\usepackage{graphicx}
\usepackage{float}
\usepackage{url}

\definecolor{Gray}{gray}{0.85}
\newcolumntype{a}{>{\columncolor{Gray}}c}
\newcolumntype{b}{>{\columncolor{white}}c}

\aclfinalcopy 

\title{What Makes Reading Comprehension Questions Easier?}

\newcommand{\sx}[0]{\spadesuit}
\newcommand{\sy}[0]{\clubsuit}
\newcommand{\sz}[0]{\diamondsuit}
\newcommand{\sw}[0]{\heartsuit}

\author{Saku Sugawara$^\sx$, Kentaro Inui$^{\sy\sz}$, Satoshi Sekine$^\sz$, Akiko Aizawa$^{\sw\sx}$ \\
  $^\sx$The University of Tokyo, $^\sy$Tohoku University,\\
  $^\sz$RIKEN AIP, $^\sw$National Institute of Informatics \\
  {\tt saku@is.s.u-tokyo.ac.jp, inui@ecei.tohoku.ac.jp,} \\
  {\tt satoshi.sekine@riken.jp, aizawa@nii.ac.jp} \\}

\date{}

\begin{document}
\maketitle
\begin{abstract}
  A challenge in creating a dataset for machine reading comprehension (MRC) is to collect questions that require a sophisticated understanding of language to answer beyond using superficial cues.
  In this work, we investigate what makes questions easier across recent 12 MRC datasets with three question styles (answer extraction, description, and multiple choice).
We propose to employ simple heuristics to split each dataset into \emph{easy} and \emph{hard} subsets and examine the performance of two baseline models for each of the subsets.
We then manually annotate questions sampled from each subset with both validity and requisite reasoning skills to investigate which skills explain the difference between easy and hard questions.
From this study, we observed that (i) the baseline performances for the hard subsets remarkably degrade compared to those of entire datasets, (ii) hard questions require knowledge inference and multiple-sentence reasoning in comparison with easy questions, and (iii) multiple-choice questions tend to require a broader range of reasoning skills than answer extraction and description questions.
  These results suggest that one might overestimate recent advances in MRC.
\end{abstract}

\section{Introduction}
\label{sec:intro}

Evaluating natural language understanding (NLU) systems is a long-established problem in AI \cite{levesque2014behaviour}. One approach to doing so is the machine reading comprehension (MRC) task, in which a system answers questions about given texts \cite{hirschman1999deep}. Although recent studies have made advances \cite{yu2018qanet}, it is still unclear to what precise extent questions require understanding of texts \cite{jia2017adversarial}.

In this study, we examine MRC datasets and discuss what is needed to create datasets suitable for detailed testing of NLU. Our motivation originates from studies that demonstrated unintended biases in the sourcing of other NLU tasks, in which questions contain simple patterns and systems can recognize these patterns to answer them \cite{gururangan2018annotation,mostafazadeh2017lsdsem}.

\begin{figure}[tb]
  \centering \small
  \newcommand{\highlight}[1]{\textcolor{blue}{\textbf{#1}}}
  \newcommand{\highlightx}[1]{\textcolor{red}{#1}}
  \fbox{%
    \parbox{0.9\linewidth}{%
      \textbf{Article:} Spectre (2015 film) on Wikipedia\\
      \textbf{Context:} ($s_1$) In \highlightx{\textit{November 2014}}, \highlight{Sony} \highlight{Pictures} Entertainment was targeted by \highlight{hackers} who released details of confidential \highlight{e-mails} between \highlight{Sony} executives regarding [...]. 
      ($s_2$) Included within these were several memos relating to the production [...]. 
      ($s_3$) Eon Productions later issued a statement [...]. \\  
      \textbf{Question:} \highlightx{When $_{(k=1)}$} did hackers get into the Sony Pictures e-mail system? \\
      \textbf{Prediction for the full question:} \textit{November 2014} \\
      \textbf{Prediction for the $k=1$ question:} \textit{November 2014} \\
      \textbf{Uni-gram overlaps between $s_i$ and the question:} \\
      $s_1$: \textcolor{blue}{5}, $s_2$: 0, $s_3$: 0 
    }
  }
  \caption{Example from the SQuAD dataset \cite{rajpurkar2016squad}. The baseline system can answer the token-limited question and, even if there are other candidate answers, it can easily attend to the answer-contained sentence ($s_1$) by watching word overlaps.}
  \label{fig:intro}
\end{figure}

We conjecture that a situation similar to this occurs in MRC datasets. Consider the question shown in Figure \ref{fig:intro}, for example.
Although the question, starting with \textit{when}, requires an answer that is expressed as a moment in time, there is only one such expression (i.e., \textit{November 2014}) in the given text (we refer to the text as the \textit{context}).
This means that the question has only a single candidate answer. The system can solve it merely by \textit{recognizing the entity type} required by \textit{when}. 
In addition to this, even if another expression of time appears in other sentences, there is only one sentence (i.e., $s_1$) that appears to be related to the question, and thus the system can easily determine the correct answer by \textit{attention}, that is, by matching the words appearing both in the context and the question.
Therefore, this kind of question does not require a complex understanding of language---e.g., multiple-sentence reasoning, which is known as a more challenging task \cite{richardson2013MCTest}.

In Section \ref{sec:heuristics}, we define two heuristics, namely \textit{entity-type recognition} and \textit{attention}.
Specifically, we analyze the differences in the performance of baseline systems for the following two configurations: (i) questions answerable or unanswerable with the first $k$ tokens; and (ii) questions whose correct answer appears or does not appear in the context sentence that is most similar to the question (henceforth referred to as \textit{the most similar sentence}).
Although similar heuristics are proposed by \newcite{weissenborn2017making}, ours are utilized for question filtering, rather than system development;
Using these simple heuristics, we split each dataset into \textit{easy} and \textit{hard} subsets for further investigation on the baseline performance.

After conducting the experiments, in Section \ref{sec:annot}, we analyze the following two points.
First, we consider which questions are valid for testing, i.e., reasonably solvable.
Second, we consider what reasoning skills are required and whether this exposes any differences among the subsets.
To investigate these two concerns, we manually annotate sample questions from each subset in terms of validity and required reasoning skills, such as word matching, knowledge inference, and multiple sentence reasoning.

We examine 12 recently proposed MRC datasets (Table \ref{tbl:datasets}), which include answer extraction, description, and multiple-choice styles. We also observe differences based on these styles.
For our baselines, we use two neural-based systems, namely, the Bidirectional Attention Flow \cite{seo2017bidirectional}
and the Gated-Attention Reader \cite{dhingra2017gated}.

In Section \ref{sec:discussion}, we describe the advantages and disadvantages of different question styles with regard to evaluating of NLU systems, and we interpret our heuristics for constructing realistic MRC datasets.

Our contributions are as follows:
\begin{itemize}[leftmargin=1em] \setlength{\parskip}{0pt}
\item This study is the first large-scale investigation across recent 12 MRC datasets with three question styles. 
\item We propose to employ simple heuristics to split each dataset into \emph{easy} and \emph{hard} subsets and examine the performance of two baseline models for each of the subsets.
\item We manually annotate questions sampled from each subset with both validity and requisite reasoning skills to investigate which skills explain the difference between easy and hard questions. 
\end{itemize}
We observed the following:
\begin{itemize}[leftmargin=1em] \setlength{\parskip}{0pt}
\item The baseline performances for the hard subsets remarkably degrade compared to those of entire datasets.
\item Our annotation study shows that hard questions require knowledge inference and multiple-sentence reasoning in comparison with easy questions.
\item Compared to questions with answer extraction and description styles, multiple-choice questions tend to require a broader range of reasoning skills while exhibiting answerability, multiple answer candidates, and unambiguity.
\end{itemize}

These findings suggest that one might overestimate recent advances in MRC systems.
They also emphasize the importance of considering simple answer-seeking heuristics when sourcing questions, in that a dataset could be easily biased unless such heuristics are employed.\footnote{All scripts used in this study, along with the subsets of the datasets and the annotation results, are available at \url{https://github.com/Alab-NII/mrc-heuristics}.}

\section{Examined Datasets and Baselines}
\label{sec:preparation}

\subsection{Datasets}

\begin{table}[t]
  \centering \small
  \fontsize{10pt}{12.0pt}\selectfont
  \begin{tabular}{l} \toprule
    \textbf{Answer extraction} (select a context span)\\
    \hphantom{0}1. SQuAD (v1.1) \cite{rajpurkar2016squad} \\
    \hphantom{0}2. AddSent \cite{jia2017adversarial} \\
    \hphantom{0}3. NewsQA \cite{trischler2017newsqa} \\
    \hphantom{0}4. TriviaQA (Wikipedia set) \cite{joshi2017triviaqa} \\
    \hphantom{0}5. QAngaroo (WikiHop) \cite{welbl2018constructing} \\ \midrule
    \textbf{Description} (generate a free-form answer) \\
    \hphantom{0}6. MS MARCO (v2) \cite{nguyen2016msmarco} \\
    \hphantom{0}7. NarrativeQA (summary) \cite{kocisky2018narrativeqa} \\ \midrule
    \textbf{Multiple choice} (choose from multiple options)\\
    \hphantom{0}8. MCTest (160 + 500) \cite{richardson2013MCTest} \\
    \hphantom{0}9. RACE (middle + high) \cite{lai2017race} \\
    10. MCScript \cite{ostermann2018mcscript} \\
    11. ARC Easy (ARC-E) \cite{clark2018think} \\
    12. ARC Challenge (ARC-C) \cite{clark2018think} \\
    \bottomrule
  \end{tabular}
  \caption{Examined datasets.}
  \label{tbl:datasets}
\end{table}

We analyzed 12 MRC datasets with three question styles: answer extraction, description, and multiple choice (Table \ref{tbl:datasets}).
Our aim was to select datasets varying in terms of corpus genre, context length, and question sourcing methods.\footnote{Because the ARC Easy and Challenge were collected using different methods, we treated them as different datasets (see \newcite{clark2018think} for further details).}
Other datasets that are not covered in our study, but can be analyzed using the same method, include:
QA4MRE \cite{sutcliffe2013QA4MRE}, CNN/Daily Mail \cite{hermann2015teaching}, Children's Book Test \cite{hill2015goldilocks}, bAbI \cite{weston2015bAbI}, WikiReading \cite{hewlett2016wikireading}, LAMBADA \cite{paperno2016lambada}, Who-did-What \cite{onishi2016who}, ProPara \cite{dalvi2018tracking}, MultiRC \cite{khashabi2018looking}, CliCR \cite{suster2018clicr}, SQuAD (v2.0) \cite{rajpurkar2018know}, and DuoRC \cite{saha2018duorc}.

\subsection{Baseline Systems}
\label{sec:systems}

We employed the following two widely used baselines.

\textbf{Bidirectional Attention Flow (BiDAF)} \cite{seo2017bidirectional} was used for the answer extraction and description datasets.
BiDAF models bi-directional attention between the context and question.
It achieved state-of-the-art performance on the SQuAD dataset.

\textbf{Gated-Attentive Reader (GA)} \cite{dhingra2017gated} was used for the multiple-choice datasets.
GA has a multi-hop architecture with an attention mechanism.
It achieved state-of-the-art-performance on the CNN/Daily Mail and Who-did-What datasets.

\textbf{Why we used different baseline systems:} The multiple-choice style can be transformed to answer extraction, as mentioned in \newcite{clark2018think}. However, in some datasets, many questions have no textual overlap to determine the correct answer span in the context. Therefore, in order to avoid underestimating the baseline performance of those datasets, we used the GA system which is applicable to multiple choice questions.

We scored the performance using exact match (EM)/F1 \cite{rajpurkar2016squad}, Rouge-L \cite{lin2004rouge}, and accuracy for the answer extraction, description, and multiple-choice datasets, respectively (henceforth, we refer to these collectively as the \textit{score}, for simplicity). For the description datasets, we determined in advance the answer span of the context that gives the highest Rouge-L score to the human-generated gold answer. We computed the Rouge-L score between the predicted span and the gold answer.\footnote{We used the official evaluation scripts of SQuAD and MS MARCO to compute the EM/F1 and Rouge-L, respectively.}

\textbf{Reproduction of the baseline performance:}
We used the same architecture as the official baseline systems unless specified otherwise.
All systems were trained on the training set and tested on the development/test set of each dataset, and we used different hyperparameters for each dataset according to characteristics such as the context length (see Appendix A for details).
We show the baseline performance of both the official results and those from our implementations in Tables \ref{tbl:res1} and \ref{tbl:res2}.
Our implementations outperformed or showed comparable performance to the official baseline on most datasets. However, in TriviaQA, MCTest, RACE, and ARC-E, our baseline performance did not reach that of the official baseline, due to differences in architecture or the absence of reported hyperparameters in the literature.

\section{Two Filtering Heuristics}
\label{sec:heuristics}

The first goal of this paper is to determine whether there are unintended biases of the kind exposed in Figure \ref{fig:intro} in MRC datasets.
We examined the influence of the two filtering heuristics: (i) entity type recognition (Section \ref{sec:qk}) and (ii) attention (Section \ref{sec:simcon}).
We then investigated the performance of the baseline systems on the questions filtered by the defined heuristics (Section \ref{sec:reeval}).

\begin{table*}[t]
  \centering \small 
  \def\arraystretch{1.1}
  \setlength{\tabcolsep}{5pt}
  \newcommand{\intermidrule}{\cmidrule{2-9}}
  \newcommand{\linestack}[1]{    \def\arraystretch{0.8}\begin{tabular}[c]{@{}c@{}} #1 \end{tabular}}
  \newcommand{\linestackx}[1]{    \def\arraystretch{0.1}\begin{tabular}[c]{@{}c@{}} #1 \end{tabular}}
  \begin{tabular}{clccccccc} \toprule
    & Dataset & SQuAD & AddSent & NewsQA & \shortstack{TriviaQA}  & QAngaroo & MARCO & NarraQA \\ 
    \midrule
    \parbox[t]{2mm}{\multirow{10}{*}{\rotatebox[origin=c]{90}{Statistics}}} & Question style (metrics) & \multicolumn{5}{c}{answer extraction (exact match / F1)} & \multicolumn{2}{|c}{description (Rouge-L)}     \\ 
    & Question sourcing & \linestack{reading \\ context} & \linestack{reading \\ context} & \linestack{reading \\ headline} & \linestack{trivia\\/ quiz} & \linestackx{chaining \\ \hphantom{$^1$}knowledge$^1$} & \linestackx{search \\ \hphantom{$^1$}query$^1$} & \linestack{reading \\ summary}     \\ 
    & Context genre & Wikipedia & Wikipedia & news & Wikipedia & Wikipedia & web & moviescript\\ 
    \intermidrule
    & Split examined     &   dev  & dev & test  & \hphantom{$^2$}dev$^2$  &    dev  & dev   & test \\
    & \# questions       &   10570 & 3560     & 5126  & 430        &  5129  & \hphantom{$^3$}55578$^3$ & 10557 \\
    & Avg. \# context tokens &   150.1 & 163.3    & 698.8 & 783.4      & 1545.5 & 625.7  & 664.5\\
    & Avg. \# question tokens &    11.8 &  12.3    & 8.0   & 19.0       &    3.6 & 6.1   & 9.9  \\
    & Avg. \# sents in context   &  5.2 &     5.8    & 30.3   & 28.5      &   57.2 & 31.5   & 27.6 \\
    \midrule
    \parbox[t]{2mm}{\multirow{13}{*}{\rotatebox[origin=c]{90}{Baseline performance}}} & Official baseline &               67.7/77.3 & 28.2/34.3 & 34.1/48.2 & 47.5/53.7 & 42.9/-    & \hphantom{$^4$}17.7$^4$   & 36.30 \\
    & \textbf{Our BiDAF baseline} & \textbf{          67.9/77.2} & \textbf{42.6/50.4} & \textbf{40.2/56.4} & \textbf{44.0/49.3} & \textbf{43.8/49.3} & \textbf{\hphantom{$^3$}36.42$^3$  } & \textbf{43.66} \\
    \intermidrule
    & Q first tokens ($k$=4)        & 30.7/44.6 & 19.2/29.7 & 30.4/44.4 & 20.5/25.0 & 43.6/49.1 & 32.61   & 25.23 \\
& \hphantom{Q first tokens} ($k$=2) & 14.0/25.0 &  9.4/17.8 & 19.4/30.3 & 14.4/18.5 & 42.6/48.0 & 25.13   & 13.00 \\
    & \hphantom{Q first tokens} ($k$=1) &  7.0/14.9 &  4.2/10.6 & 13.5/23.8 & 8.6/12.5  & 42.0/47.5 & 21.67   &  8.45 \\
    & \% of \# Q ($\geq$0.5 for $k$=2)         & 22.4  & 15.8      & 29.7      & 20.0      & 49.8      & 17.9  & 10.3 \\ 
    \intermidrule
          & Ans in sim sent & 71.4/80.6 & 50.2/58.2 & 42.9/59.7 & 58.0/65.1 & 41.7/49.2 & 38.96 & 45.17 \\
& \hphantom{And} only with sim sent  & 73.3/82.8 & 71.4/81.1 & 52.8/70.9 & 64.8/72.7 & 66.7/74.2 & 45.30 & 58.56 \\
                & Ans not in sim sent & 56.6/66.4 & 28.1/35.5 & 37.8/53.5 & 40.4/45.2 & 43.9/49.3 & 35.84 & 41.99 \\
    & \% of \# Q (ans in sim)           & 76.3  & 65.7      & 46.3      & 20.5      & 4.2       & 18.6  & 52.6 \\
    \intermidrule
    & \textbf{\textit{Hard} subset}                & \textbf{38.7/45.2} & \textbf{18.2/23.4} & \textbf{27.9/40.9} & \textbf{30.0/32.5} & \textbf{2.3/2.6  } & \textbf{15.42 } & \textbf{39.61} \\
    & \textbf{\% of \textit{hard}                 } & \textbf{15.7  } & \textbf{25.4     } & \textbf{30.0     } & \textbf{59.8     } & \textbf{36.9     } & \textbf{12.5  } & \textbf{28.2} \\
    \bottomrule  
  \end{tabular}
  \caption{Statistics from the answer extraction and description datasets and their baselines.
    \textit{Dev} represents a development set. \textit{Ans in sim sent} refers to questions whose answer appears in the sentence that is most similar to the question.
    $^1$The questions are not complete sentences and may start with more specific words than interrogatives.
    $^2$Verified set.
    $^3$No answer questions were removed.
    $^4$The Passage Ranking model \cite{nguyen2016msmarco}.}
  \label{tbl:res1}
\end{table*}

\begin{table}[t]
  \centering 
  \fontsize{8pt}{8pt}\selectfont
  \def\arraystretch{1.2}
  \newcommand{\intermidrule}{\cmidrule{2-7}}
  \newcommand{\linestack}[1]{\def\arraystretch{0.9}\begin{tabular}[c]{@{}c@{}} #1 \end{tabular}}
  \setlength{\tabcolsep}{1.8pt}
  \begin{tabular}{clccccc} \toprule
    & Dataset & MCTest & RACE & MCScript & ARC-E & ARC-C  \\
    \midrule
    \parbox[t]{2mm}{\multirow{10}{*}{\rotatebox[origin=c]{90}{Statistics}}} & Style (metrics) & \multicolumn{5}{c}{multiple choice (accuracy)} \\
    & Q sourcing & \linestack{reading \\ context} & \linestack{English \\ exam} & \linestack{script \\ scenario} & \multicolumn{2}{c}{science exam} \\
    & Genre & narrative & various & narrative & \multicolumn{2}{c}{textbook} \\
    \intermidrule
    & Split examined     &   test  & test & dev  & dev & dev \\
    & \# questions       &  840   & 4934 & 1411 & 2376 & 1171  \\
    & Avg. \# C tokens   & 249.9  & 339.3 & 195.2 & 142.0 & 138.3  \\
    & Avg. \# Q tokens   &  9.4   & 11.5 & 7.8  & 21.8 & 25.4  \\
    & Avg. \# sents      & 18.4   & 17.9 & 11.5 & 8.1 & 8.2  \\
    \midrule
    \parbox[t]{2mm}{\multirow{12}{*}{\rotatebox[origin=c]{90}{Baseline performance}}}
    & Random                          & 25.0      & 25.0      & 50.0      & 25.0      & 25.0 \\
    & Official baseline               & \hphantom{$^1$}43.2$^1$      & 44.1      & 72.0      & \hphantom{$^2$}62.6$^2$      & \hphantom{$^2$}20.3$^2$ \\
    & \textbf{Our GA baseline                   } & \textbf{34.3     } & \textbf{42.7     } & \textbf{75.5     } & \textbf{43.9     } & \textbf{30.1} \\
    \intermidrule
    & Q tokens ($k$=4)                 & 36.1  & 38.4      & 73.7      & 38.8      & 30.6 \\
& \hphantom{Q tokens} ($k$=2)          & 33.9      & 37.7      & 71.1      & 37.0      & 29.0 \\
    & \hphantom{Q tokens} ($k$=1)      & 34.9  & 36.4      & 70.9      & 35.3      & 28.6 \\
    \intermidrule
          & Ans in sim sent          & 33.1      & 40.8      & 74.0      & 47.5      & 31.6 \\
          &   \hphantom{ans }only w/ sim                   & 32.4      & 40.4      & 74.4      & 48.5      & 28.9 \\
               & Ans not in sim        & 34.9      & 43.3      & 75.8      & 40.4      & 29.4 \\
             & \% of \# Q (in sim)      & 33.5      & 23.2      & 17.7      & 48.7      & 34.8 \\
    \intermidrule
    & \textbf{\textit{Hard} subset               } & \textbf{ 4.3     } & \textbf{23.5     } & \textbf{28.7     } & \textbf{20.6     } & \textbf{15.6} \\
    & \textbf{\% of \textit{hard}             } & \textbf{62.4     } & \textbf{58.8     } & \textbf{27.1     } & \textbf{53.9     } & \textbf{66.4} \\
    \bottomrule
  \end{tabular}
  \caption{Statistics from the multiple-choice datasets and their baselines. $^1$The Attentive Reader \cite{hermann2015teaching} from \newcite{yin2016HABCNN}. $^2$An information retrieval system from \newcite{clark2018think}.}
  \label{tbl:res2}
\end{table}

\subsection{Entity Type-based Heuristic}
\label{sec:qk}

The aim of this heuristic was to detect questions that can be solved based on (i) the existence of a single candidate answer that is restricted by expressions such as ``wh-'' and ``how many,'' and (ii) lexical patterns that appear around the correct answer.
Because the query styles are not uniform across datasets (e.g., MARCO uses search engine queries), we could not directly use interrogatives.
Instead, we simply provided the first $k$ tokens of questions to the baseline systems.
We choose smaller values for $k$ than the (macro) average of the question length across the datasets (= 12.2 tokens).
For example, for $k=4$ of the question \textit{will I qualify for OSAP if I'm new in Canada} (excerpted from MARCO), we use \textit{will I qualify for}.
Even if the tokens do not have an interrogative, the system may recognize lexical patterns around the correct answer.
Questions that can be solved by examining these patterns were also of interest when filtering.

\textbf{Results:} The results for $k=1,2,4$ are shown in Tables \ref{tbl:res1} and \ref{tbl:res2}.
In addition, to know the exact ratio of the questions that are solved rather than the scores for the answer extraction and description styles, we counted questions with $k=2$ that achieved the score $\geq$ 0.5.\footnote{We considered that this threshold is sufficient to judge that the system attends to the correct span because of the potential ambiguity of these styles (see Section \ref{sec:annot}).}
As $k$ decreased, so too did the baseline performance on all datasets in Table \ref{tbl:res1} except QAngaroo.
By contrast, in QAngaroo and the multiple-choice datasets, the performance did not degrade so strongly.
In particular, the difference between the scores on the full and $k=1$ questions in QAngaroo was 1.8.
Because the questions in QAngaroo are not complete sentences, but rather knowledge-base entries that have a blank, such as \textit{country\_of\_citizenship Henry VI of England,} this result implies that the baseline system can infer the answer merely by the first token of questions, i.e., the type of knowledge-base entry.

In most multiple-choice datasets, the $k=1$ scores were significantly higher than random-choice scores. Given that multiple-choice questions offer multiple options that are of valid entity/event types, this gap was not necessarily caused by the limited number of candidate answers, as in the case with the answer extraction datasets.
We therefore infer that, in the solved questions, incorrect options appear less than the correct option does, or they do not appear at all in the context (such questions are regarded as solvable exclusively by using the word match skill, which we analyze in Section \ref{sec:annot}).
Remarkably, though we failed to achieve a higher baseline performance, the score for complete questions in MCTest was lower than the score of the $k=1$ questions. This shows that the MCTest questions are sufficiently difficult such that it was not especially useful for the baseline system to consider the entire question statement.

\subsection{Attention-based Heuristic}
\label{sec:simcon}

Next, we examined in each dataset (i) how many questions have their correct answers in the most similar sentence, and (ii) whether there is a performance gap for such questions (i.e., whether such questions are easier than the others).

We used uni-gram overlap as a similarity measure.\footnote{Although there are other similarity measures, we used this basic measure to obtain an intuitive result.}
We counted how many times question words appear in each sentence where question words are stemmed and stopwords are dropped.
Then we checked whether the correct answer appears in the most similar sentence.
For multiple-choice datasets, we selected the text span that gives the highest Rouge-L score with the correct option as the correct answer.

\textbf{Results:} The results are shown in Tables \ref{tbl:res1} and \ref{tbl:res2}. Considering the average number of context sentences, most datasets contained a significantly high proportion of questions whose answers were in the most similar sentence.

In the answer extraction and description datasets (except QAngaroo), the baseline performance improved when the correct answer appeared in the most similar sentence and there were gaps between the performances on these questions and the others.
These gaps indicate that the dataset may lack balance for testing NLU; if these questions tend to require word matching skill exclusively, attending the other portion is useful to study more realistic NLU, e.g., common-sense reasoning and discourse understanding. Therefore, we investigated whether these questions merely require word matching (see Section \ref{sec:annot}).

On the other hand, in the first three multiple-choice datasets, the performance differences were marginal or inversed. This implies that, although the baseline performance was not especially high, the difficulty of these questions for the baseline system was not affected by whether their correct answers appeared in the most similar sentence.

We further analyzed the baseline performance after removing the context and leaving only the most similar sentence. In AddSent and QAngaroo, the scores improved remarkably ($>$20 F1); from this result, we can infer that on these datasets the baseline systems are distracted by other sentences in the context. This observation is supported by the results from the AddSent dataset \cite{jia2017adversarial}, which contains manually-injected distracting sentences (i.e., adversarial examples).

\subsection{Performance on \textit{Hard} Subsets}
\label{sec:reeval}

In the previous two sections, we observed that in the examined datasets (i) some questions were solved by the baseline systems merely with the first $k$ tokens and/or (ii) the baseline performances increased for questions whose answers were in the most similar sentence.
Because we were concerned that these two become dominant factors in measuring the baseline performance using the datasets,
we split each development/test set into \textit{easy} and \textit{hard} subsets for further investigation.

\textbf{\textit{Hard} subsets:} A \textit{hard} subset comprised questions (i) whose score is not positive when $k=2$ \textit{and} (ii) whose correct answer does not appear in the most similar sentence. The easy subsets comprised the remaining questions.
We aimed to investigate the gap of performance values between the \textit{easy} and \textit{hard} subsets. If the gap is large, the dataset may be strongly biased toward questions that are solved by recognizing entity types or lexical patterns and may not be suitable for measuring the system's ability for complex reasoning.


\textbf{Results and clarification:}
The bottom row in Tables \ref{tbl:res1} and \ref{tbl:res2} shows that the baseline performances on the \textit{hard} subset remarkably decreased in almost all examined datasets.
These results reveal that we may overestimate the ability of the baseline systems perceived previously.
However, we clarify that our intention is not to remove questions solved or mitigated by our defined heuristics to create a new \textit{hard} subset, since this may generate new biases as indicated in \newcite{gururangan2018annotation}.
Rather, we would like to emphasize the importance of the defined heuristics when sourcing questions. Indeed, ill attention to these heuristics can lead to unintended biases.

\section{Annotating Question Validity and Required Skills}
\label{sec:annot}

\subsection{Annotation Specifications}
\label{sec:annotobjectives}

\textbf{Objectives:} To complement the observations in the previous sections, we annotated sampled questions from each subset of the datasets.
Our motivation can be summarized as follows:
(i) How many questions are valid in each dataset? That is, the \textit{hard} questions may not in fact be hard, but just unsolvable, as indicated in \newcite{chen2016thorough}.
(ii) What kinds of reasoning skills explain \textit{easy}/\textit{hard} questions?
(iii) Are there any differences among the datasets and the question styles?

We annotated the minimum skills required to choose the correct answer among other candidates. We assumed that the solver knows what type of entity or event is entailed by the question.

\textbf{Annotation labels:} Our annotation labels (Table \ref{tbl:annotlabels}) were inspired by previous work such as \newcite{chen2016thorough}, \newcite{trischler2017newsqa}, and \newcite{lai2017race}.
The major modifications were twofold: (i) detailed question validity, including a number of reasonable candidate answers and answer ambiguity; and (ii) posing multiple-sentence reasoning as a skill compatible with other skills.

\begin{table}[t]
  \centering \fontsize{10pt}{11.5pt}\selectfont
  \setlength{\tabcolsep}{2.5pt}
  \def\arraystretch{1.05}
  \begin{tabular}{lp{0.9\linewidth}} \toprule
    \multicolumn{2}{l}{\textbf{Validity}} \\
    1. & \textit{Unsolvable} -- the context coupled with the question does not reasonably give the answer. \\
    2. & \textit{Single candidate} -- the question does not have multiple candidate answers. \\
    3. & \textit{Ambiguous} -- the question does not have a unique, decidable answer, or, multiple possible answers are not covered by the gold answers. \\
    \midrule \multicolumn{2}{l}{\textbf{Reasoning skill}} \\
    4. & \textit{Word matching} -- matching the context and question words. \\
    5. & \textit{Paraphrasing} -- using lexical and grammatical knowledge. \\
    6. & \textit{Knowledge} -- inference using commonsense and/or world knowledge. \\
    7. & \textit{Meta/Whole} -- understanding of meta terms such as the ``author'' and ``writer,'' and comprehending the general context. \\
    8. & \textit{Math/Logic} -- using mathematical and logical knowledge. This includes multiple-choice questions that ask ``which option is not true.'' \\
    \midrule \multicolumn{2}{l}{\textbf{Multiple-sentence reasoning}} \\
    9. & (i) coreference (ii) causal relation (iii) spacial-temporal relations (iv) none -- gathering cues from multiple sentences/clauses. \\
  \bottomrule
  \end{tabular}
  \caption{Annotation labels. One of reasoning skills is annotated with the questions that are ``no'' in all validity labels. Multiple sentence reasoning is independent of reasoning skills and annotated with all valid questions.}
  \label{tbl:annotlabels}
\end{table}

Indeed there are other classifications of reasoning types.
For instance, \newcite{lai2017race} defined five reasoning types, including \textit{attitude analysis} and \textit{whole-picture reasoning}.
We incorporated them into the \textit{knowledge} and \textit{meta/whole} classes.
\newcite{clark2018think} proposed detailed knowledge and reasoning types, but these were specific to science exams, and thus omitted from our study.

Independent of the reasoning types above, we checked whether the question required multiple-sentence reasoning to answer the questions.
As another modification, we extended the notion of ``sentence'' in our annotation and considered a subordinate clause as a sentence.
This modification was intended to deal with the internal complexity of a sentence with multiple clauses, which can also render a question difficult.

\textbf{Settings:}
For each subset of the datasets, 30 questions were annotated.
Therefore we obtained annotations for $30 \times 2 \times 12 = 720$
questions.
The annotation was performed by the authors.
The annotator was given the context, question, and candidate answers for multiple-choice questions, along with the correct answer. To reduce bias, the annotator did not know which \textit{easy} or \textit{hard} subset the questions were in, and was not told the predictions and scores of the respective baseline systems.

\subsection{Annotation Results}
\label{sec:annotres}

\begin{table*}[t]
  \centering \small
  \def\arraystretch{1.0}
  \newcommand{\intermidrule}{\cmidrule{2-16}}  
  \setlength{\tabcolsep}{4pt}
  \setlength{\aboverulesep}{0pt}
  \setlength{\belowrulesep}{0pt}
  \setlength{\extrarowheight}{.75ex}
  \begin{tabular}{clcacacacacacaca} \toprule
    & Dataset & \multicolumn{2}{c}{SQuAD} & \multicolumn{2}{c}{AddSent} & \multicolumn{2}{c}{NewsQA} & \multicolumn{2}{c}{TriviaQA} & \multicolumn{2}{c}{QAngaroo} & \multicolumn{2}{c}{MARCO} & \multicolumn{2}{c}{NarraQA} \\
    \midrule
    & Subset & easy & hard & easy & hard & easy & hard & easy & hard & easy & hard & easy & hard & easy & hard \\ 
    \midrule & F1/Rouge-L	 & 80.9 & 37.6 & 61.5 & 29.5 & 52.7 & 30.3 & 70.6 & 33.4 & 71.1 & 3.5 & 49.4 & 21.5 & 54.9 & 51.2 \\ 
    \midrule 
        \parbox[t]{2mm}{\multirow{4}{*}{\rotatebox[origin=c]{90}{Validity}}} & Unsolvable	 & 0.0 & 0.0 & 0.0 & 0.0 & 0.0 & 6.7 & 16.7 & 16.7 & 33.3 & 43.3 & 0.0 & 0.0 & 0.0 & 0.0 \\ 
    & Single cand.	 & 23.3 & 10.0 & 6.7 & 3.3 & 10.0 & 3.3 & 3.3 & 6.7 & 6.7 & 3.3 & 0.0 & 0.0 & 6.7 & 0.0 \\ 
    & Ambiguous	 & 3.3 & 13.3 & 3.3 & 13.3 & 43.3 & 30.0 & 13.3 & 13.3 & 13.3 & 20.0 & 6.7 & 3.3 & 0.0 & 0.0 \\ 
        & Valid	 & 73.3 & 76.7 & 90.0 & 83.3 & 46.7 & 60.0 & 66.7 & 63.3 & 46.7 & 33.3 & 93.3 & 96.7 & 93.3 & 100.0 \\
        \midrule 
    \parbox[t]{2mm}{\multirow{5}{*}{\rotatebox[origin=c]{90}{Skill}}} &  Word match & 59.1 & 21.7 & 55.6 & 24.0 & 42.9 & 66.7 & 45.0 & 26.3 & 35.7 & 20.0 & 89.3 & 44.8 & 46.4 & 43.3 \\ 
    & Paraphrasing	 & 18.2 & 26.1 & 11.1 & 36.0 & 21.4 & 11.1 & 5.0 & 10.5 & 7.1 & 20.0 & 0.0 & 10.3 & 25.0 & 20.0 \\ 
    & Knowledge	 & 22.7 & 47.8 & 33.3 & 40.0 & 35.7 & 22.2 & 50.0 & 63.2 & 57.1 & 60.0 & 10.7 & 44.8 & 28.6 & 33.3 \\ 
    & Meta/Whole	 & 0.0 & 0.0 & 0.0 & 0.0 & 0.0 & 0.0 & 0.0 & 0.0 & 0.0 & 0.0 & 0.0 & 0.0 & 0.0 & 3.3 \\ 
    & Math/Logic	 & 0.0 & 4.3 & 0.0 & 0.0 & 0.0 & 0.0 & 0.0 & 0.0 & 0.0 & 0.0 & 0.0 & 0.0 & 0.0 & 0.0 \\ \midrule
    \parbox[t]{2mm}{\multirow{4}{*}{\rotatebox[origin=c]{90}{Relation}}} & Multi sent.	 & 22.7 & 17.4 & 25.9 & 36.0 & 35.7 & 16.7 & 35.0 & 36.8 & 57.1 & 80.0 & 7.1 & 13.8 & 28.6 & 46.7 \\
    \intermidrule
    & Coreference	 & 18.2 & 17.4 & 14.8 & 32.0 & 21.4 & 16.7 & 35.0 & 31.6 & 50.0 & 50.0 & 7.1 & 13.8 & 14.3 & 33.3 \\
    & Causal	 & 0.0 & 0.0 & 3.7 & 4.0 & 0.0 & 0.0 & 0.0 & 0.0 & 0.0 & 0.0 & 0.0 & 0.0 & 14.3 & 6.7 \\ 
    & Space/Temp.	 & 4.5 & 0.0 & 7.4 & 0.0 & 14.3 & 0.0 & 0.0 & 5.3 & 7.1 & 30.0 & 0.0 & 0.0 & 0.0 & 6.7
  \\ \bottomrule \end{tabular}
  \caption{Annotation results for the answer extraction and description datasets.}
  \label{tbl:annotae}
\end{table*}

\begin{table*}[!htb]
  \centering \small
  \def\arraystretch{1.0}
  \newcommand{\intermidrule}{\cmidrule{2-12}}
  \setlength{\tabcolsep}{4pt}
  \setlength{\aboverulesep}{0pt}
  \setlength{\belowrulesep}{0pt}
  \setlength{\extrarowheight}{.75ex}
  \begin{tabular}{clcacacacaca} \toprule
    & Dataset & \multicolumn{2}{c}{MCTest} & \multicolumn{2}{c}{RACE} & \multicolumn{2}{c}{MCScript} & \multicolumn{2}{c}{ARC-E} & \multicolumn{2}{c}{ARC-C} \\
    \midrule
   & Subset & easy & hard & easy & hard & easy & hard & easy & hard & easy & hard \\ 
   \midrule
   & Accuracy & 83.3 & 13.3 & 76.7 & 30.0 & 93.3 & 26.7 & 60.0 & 16.7 & 43.3 & 10.0 \\
   \midrule
   \parbox[t]{2mm}{\multirow{4}{*}{\rotatebox[origin=c]{90}{Validity}}}  & Unsolvable	 & 0.0 & 0.0 & 0.0 & 0.0 & 0.0 & 0.0 & 3.3 & 30.0 & 46.7 & 33.3 \\ 
 & Single cand. & 0.0 & 0.0 & 0.0 & 0.0 & 0.0 & 0.0 & 0.0 & 0.0 & 0.0 & 0.0 \\ 
 & Ambiguous	 & 0.0 & 0.0 & 3.3 & 0.0 & 0.0 & 0.0 & 3.3 & 0.0 & 3.3 & 3.3 \\ 
 & Valid	 & 100.0 & 100.0 & 96.7 & 100.0 & 100.0 & 100.0 & 93.3 & 70.0 & 50.0 & 63.3 \\
 \midrule
\parbox[t]{2mm}{\multirow{5}{*}{\rotatebox[origin=c]{90}{Skill}}} & Word match	 & 56.7 & 46.7 & 17.2 & 6.7 & 36.7 & 46.7 & 71.4 & 52.4 & 33.3 & 15.8 \\ 
& Paraphrasing	 & 6.7 & 10.0 & 13.8 & 6.7 & 20.0 & 6.7 & 14.3 & 19.0 & 20.0 & 31.6 \\ 
 & Knowledge	 & 30.0 & 26.7 & 34.5 & 43.3 & 20.0 & 36.7 & 14.3 & 23.8 & 40.0 & 42.1 \\ 
& Meta/Whole	 & 3.3 & 3.3 & 31.0 & 33.3 & 20.0 & 10.0 & 0.0 & 0.0 & 0.0 & 0.0 \\ 
& Math/Logic	 & 3.3 & 13.3 & 3.4 & 10.0 & 3.3 & 0.0 & 0.0 & 4.8 & 6.7 & 10.5 \\ 
\midrule
\parbox[t]{2mm}{\multirow{4}{*}{\rotatebox[origin=c]{90}{Relation}}} & Multi sent.	 & 46.7 & 73.3 & 58.6 & 76.7 & 0.0 & 30.0 & 7.1 & 14.3 & 0.0 & 10.5 \\
\intermidrule
& Coreference	 & 33.3 & 56.7 & 44.8 & 60.0 & 0.0 & 16.7 & 7.1 & 9.5 & 0.0 & 0.0 \\ 
    & Causal	 & 6.7 & 6.7 & 3.4 & 13.3 & 0.0 & 3.3 & 0.0 & 0.0 & 0.0 & 0.0 \\ 
& Space/Temp.	 & 6.7 & 10.0 & 10.3 & 3.3 & 0.0 & 10.0 & 0.0 & 4.8 & 0.0 & 10.5 \\
\bottomrule \end{tabular}
  \caption{Annotation results for the multiple-choice datasets.}
  \label{tbl:annotmc}
\end{table*}

\begin{table}[htb]
  \centering \small 
  \newcommand{\linestack}[1]{    \def\arraystretch{0.9}\begin{tabular}[c]{@{}c@{}} #1 \end{tabular}}
  \begin{tabular}{lcc} \toprule
    Label & $r$ & $p$ \\ \midrule
    Single cand (BiDAF) & 0.150 & 0.002 \\
    Ambiguous (BiDAF) & 0.098 & 0.044 \\
    Word matching (BiDAF) & 0.266 & 0.000 \\
    Knowledge (BiDAF) & -0.288 & 0.000 \\
    Multi sent (BiDAF) & -0.120 & 0.035 \\
    Unsolvable (GA) & -0.119 & 0.039 \\
    \bottomrule
  \end{tabular}
  \caption{Pearson's correlation coefficients ($r$) between the annotation labels and the baseline scores with $p$ $<$ 0.05.}
  \label{tbl:coeff}
\end{table}

Tables \ref{tbl:annotae} and \ref{tbl:annotmc} show the results of the annotation.

\textbf{Validity:} TriviaQA, QAngaroo, and ARCs revealed relatively high \textit{unsolvability}. This seemed to be caused by unrelatedness between the questions and their context. For example, QAngaroo's context was gathered from Wikipedia articles that are not necessarily related to the questions.\footnote{Nonetheless, it is remarkable that, even though the dataset was constructed automatically, the remaining valid \textit{hard} questions were difficult for the baseline system.}
The context passages in ARCs were curated from textbooks that may not provide sufficient information to answer the questions.\footnote{Our analysis was not intended to undermine the quality of these questions. We refer readers to \newcite{clark2018think}.}
Note that it is possible that unsolvable questions are permitted and that the system must indicate them in some datasets such as QA4MRE, NewsQA, MARCO, and SQuAD (v2.0).

For \textit{single candidate}, however, we found that few questions had only single-candidate answers. Further, there were even fewer single-candidate answers in AddSent than in SQuAD. This result supports the claim that the adversarial examples augment the number of possible candidate answers, thus degrading the baseline performance.

\begin{figure}[t]
  \centering \small
  \fbox{%
    \parbox{0.9\linewidth}{%
      \textbf{ID:} {\tiny ./cnn/stories/4ca29639845a40551a62d10212a46aec7caf3369.story-2} \\
      \textbf{Context:} [...] This plot of land is scheduled to house the permanent United Airlines Flight 93 memorial. [...] \\
      \textbf{Question:} What was the name of the flight? \\
      \textbf{Answer:} 93 \\
      \textbf{Possible answers:} United Airlines Flight 93, Flight 93
    }
  }
  \caption{Example of an \textit{ambiguous} question from NewsQA \cite{trischler2017newsqa}.}
  \label{fig:ambiguous}
\end{figure}

In our annotation, \textit{ambiguous} questions were found to be those with multiple correct spans. We show an example in Figure \ref{fig:ambiguous}.
In this case, several answers other than ``93'' are correct.
Ambiguity is an important feature, insofar as it can lead to unstable scoring in EM/F1.

The multiple-choice datasets mostly comprised valid questions, with the exception of the unsolvable questions in the ARC datasets.

\textbf{Reasoning skills:} We can see that \textit{word matching} was more important in the \textit{easy} subsets, and \textit{knowledge} was more pertinent to the \textit{hard} subsets in 10 of the 12 datasets.
These results confirm that the manner by which we split the subsets was successful at filtering questions that are relatively easy in terms of reasoning skills.
However, we did not observe this trend with \textit{paraphrasing}, which seemed difficult to distinguish from \textit{word matching} and \textit{knowledge}.
With regard to \textit{meta/whole} and \textit{math/logic}, we can see that these skills were needed less in the answer extraction and description datasets. They were more pertinent to the multiple-choice datasets.

\textbf{Multiple-sentence reasoning:} Multiple-sentence reasoning correlated more with the \textit{hard} subsets in 10 of the 12 datasets. Although NewsQA showed the inverse tendency for \textit{word matching}, \textit{knowledge}, and \textit{multiple-sentence reasoning}, we suspect that this was caused by annotation variance and filtering a large portion of ambiguous questions. For relational types, we did not see a significant trend in any particular type.

\textbf{Correlation of labels and baseline scores:} Across all examined datasets, we analyzed the correlations between the annotation labels and the scores of each baseline system in Table \ref{tbl:coeff}. In spite of the small size of the annotated samples, we derived statistically significant correlations for six labels.
These results confirm that BiDAF performs well for the \textit{word matching} questions and relatively poorly with the \textit{knowledge} questions.
By contrast, we did not observe this trend in GA.

\section{Discussion}
\label{sec:discussion}

In this section, we discuss the advantages and disadvantages of the question styles, and we interpret the defined heuristics in terms of constructing more realistic MRC datasets.

\textbf{Differences among the question styles:} The biggest advantage to the answer extraction style is its ease in generating questions, which enables us to produce large-scale datasets.
On the other hand, a disadvantage to this style is that it rarely demands \textit{meta/whole} and \textit{math/logic} skills, which can require answers not contained in the context.
Moreover, as observed in Section \ref{sec:annot}, it seems difficult to guarantee that all possible answer spans are given as the correct answers.
By contrast, the description and multiple-choice styles have the advantage that they have no such restrictions on the appearance of candidate answers \cite{kocisky2018narrativeqa,khashabi2018looking}.
Nonetheless, the description style is difficult to evaluate because the Rouge-L and BLEU scores are insufficient for testing NLU.
Whereas it is easy to evaluate the performance on multiple-choice questions,
generating multiple reasonable options requires considerable effort.

\textbf{Interpretation of our heuristics:}
When we regard the MRC task as recognizing textual entailment (RTE) \cite{dagan2006pascal}, the task requires the reader to construct one or more premises from the context and then form the most reasonable hypothesis from the question and candidate answer \cite{sachan2015learning}.
Thus, easier questions are those (i) where the reader needs to generate only one hypothesis, and (ii) where the premises directly describe the correct hypothesis. Our two heuristics can also be seen as the formalizations of these criteria. Therefore, to make questions more realistic, we need to create multiple hypotheses 
that require complex reasoning in order to be distinguished.
Moreover, the integration of premises should be complemented by external knowledge to provide sufficient information to verify the correct hypothesis. 

\section{Related Work}
\label{sec:related}

Our heuristics and annotation are motivated by \textit{unintended biases} \cite{levesque2014behaviour} and \textit{evaluation overfitting} \cite{whiteson2011protecting}, respectively.

\textbf{Unintended biases:} The MRC task tests a reading process that involves retrieving stored information and performing inferences \cite{sutcliffe2013QA4MRE}.
However, it is difficult to construct datasets that comprehensively require those skills.
As \newcite{levesque2014behaviour} discussed as a desideratum for testing AI, we should avoid creating questions that can be solved by matching patterns, using unintended biases, and selectional restrictions.
For the unintended biases, one suggestive example is the Story Cloze Test \cite{mostafazadeh2016corpus}, in which a system chooses a sentence among candidates to conclude a given paragraph of the story.
A recent attempt at this task showed that recognizing superficial features in the correct candidate is critical to achieve the state of the art \cite{schwartz2017story}.

Similarly, in MRC, \newcite{weissenborn2017making} proposed \textit{context/type matching heuristic} to develop a simple neural system.
\newcite{min2018efficient} observed that 92\% of answerable questions in SQuAD can be answered only using a single context sentence. 
In visual question answering, \newcite{agrawal2016analyzing} analyzed the behavior of models with the variable length of the first question words.
More recently, \newcite{khashabi2018looking} proposed a dataset that has questions for multi-sentence reasoning.

\textbf{Evaluation overfitting:}
The theory behind evaluating AI distinguishes between task- and skill-oriented approaches \cite{hernandez2017evaluation}.
In the task-oriented approach, we usually develop a system and test it on a specific dataset.
Sometimes the developed system lacks generality but achieves the state of the art for that specific dataset. Further, it becomes difficult to verify and explain the solution to tasks. 
The situation in which we are biased to the specific tasks is called evaluation overfitting \cite{whiteson2011protecting}.
By contrast, with the skill-oriented approach, we aim to interpret the relationships between tasks and skills.
This orientation can encourage the development of more realistic NLU systems.

As one of our goals was to investigate whether easy questions are dominant in recent datasets, it did not necessarily require a detailed classification of reasoning types.
Nonetheless, we recognize there are more fine-grained classifications of required skills for NLU. 
For example, \newcite{weston2015bAbI} defined 20 skills as a set of toy tasks.
\newcite{sugawara2017evaluation} also organized 10 prerequisite skills for MRC.
\newcite{lobue2011commonsense} and \newcite{sammons2010ask} analyzed entailment phenomena using detailed classifications in RTE.
For the ARC dataset, \newcite{boratko2018systematic} proposed knowledge and reasoning types.

\section{Conclusion}

In this study, MRC questions from 12 datasets were examined in order to determine what makes such questions easier to answer. We defined two heuristics that limit candidate answers and thereby mitigate the difficulty of questions.
Using these heuristics, the datasets were split into easy and hard subsets.
We further annotated the questions with their validity and the reasoning skills needed to answer them.
Our experiments revealed that the baseline performance degraded with the hard questions, which required knowledge inference and multiple-sentence reasoning compared to easy questions.
These results suggest that one might overestimate the ability of the baseline systems. They also emphasize the importance of analyzing and reporting the properties of new datasets when released.
One limitation of this work was the heavy cost of the annotation.
In future research, we plan to explore a method for automatically classifying reasoning types. This will enable us to evaluate systems through a detailed organization of the datasets.

\section*{Acknowledgments}

We would like to thank Rajarshi Das, Shehzaad Dhuliawala, and anonymous reviewers for their insightful comments. This work was supported by JSPS KAKENHI Grant Numbers 18H03297 and 18J12960.

\bibliography{emnlp2018}

\begin{thebibliography}{46}
\expandafter\ifx\csname natexlab\endcsname\relax\def\natexlab#1{#1}\fi

\bibitem[{Agrawal et~al.(2016)Agrawal, Batra, and
  Parikh}]{agrawal2016analyzing}
Aishwarya Agrawal, Dhruv Batra, and Devi Parikh. 2016.
\newblock Analyzing the behavior of visual question answering models.
\newblock In \emph{Proceedings of the 2016 Conference on Empirical Methods in
  Natural Language Processing}, pages 1955--1960, Austin, Texas. Association
  for Computational Linguistics.

\bibitem[{Boratko et~al.(2018)Boratko, Padigela, Mikkilineni, Yuvraj, Das,
  McCallum, Chang, Fokoue-Nkoutche, Kapanipathi, Mattei, Musa, Talamadupula,
  and Witbrock}]{boratko2018systematic}
Michael Boratko, Harshit Padigela, Divyendra Mikkilineni, Pritish Yuvraj,
  Rajarshi Das, Andrew McCallum, Maria Chang, Achille Fokoue-Nkoutche, Pavan
  Kapanipathi, Nicholas Mattei, Ryan Musa, Kartik Talamadupula, and Michael
  Witbrock. 2018.
\newblock A systematic classification of knowledge, reasoning, and context
  within the arc dataset.
\newblock In \emph{Proceedings of the Workshop on Machine Reading for Question
  Answering}, pages 60--70, Melbourne, Australia. Association for Computational
  Linguistics.

\bibitem[{Chen et~al.(2016)Chen, Bolton, and Manning}]{chen2016thorough}
Danqi Chen, Jason Bolton, and Christopher~D. Manning. 2016.
\newblock A thorough examination of the {CNN}/{D}aily {M}ail reading
  comprehension task.
\newblock In \emph{Proceedings of the 54th Annual Meeting of the Association
  for Computational Linguistics}, pages 2358--2367. Association for
  Computational Linguistics.

\bibitem[{Clark et~al.(2018)Clark, Cowhey, Etzioni, Khot, Sabharwal, Schoenick,
  and Tafjord}]{clark2018think}
Peter Clark, Isaac Cowhey, Oren Etzioni, Tushar Khot, Ashish Sabharwal, Carissa
  Schoenick, and Oyvind Tafjord. 2018.
\newblock Think you have solved question answering? {Try} {ARC}, the {AI2}
  reasoning challenge.
\newblock \emph{CoRR}, abs/1803.05457.

\bibitem[{Dagan et~al.(2006)Dagan, Glickman, and Magnini}]{dagan2006pascal}
Ido Dagan, Oren Glickman, and Bernardo Magnini. 2006.
\newblock The {PASCAL} recognising textual entailment challenge.
\newblock In \emph{Machine learning challenges. evaluating predictive
  uncertainty, visual object classification, and recognising tectual
  entailment}, pages 177--190. Springer.

\bibitem[{Dalvi et~al.(2018)Dalvi, Huang, Tandon, Yih, and
  Clark}]{dalvi2018tracking}
Bhavana Dalvi, Lifu Huang, Niket Tandon, Wen-tau Yih, and Peter Clark. 2018.
\newblock Tracking state changes in procedural text: a challenge dataset and
  models for process paragraph comprehension.
\newblock In \emph{Proceedings of the 2018 Conference of the North American
  Chapter of the Association for Computational Linguistics: Human Language
  Technologies, Volume 1 (Long Papers)}, pages 1595--1604. Association for
  Computational Linguistics.

\bibitem[{Dhingra et~al.(2017)Dhingra, Liu, Yang, Cohen, and
  Salakhutdinov}]{dhingra2017gated}
Bhuwan Dhingra, Hanxiao Liu, Zhilin Yang, William Cohen, and Ruslan
  Salakhutdinov. 2017.
\newblock Gated-attention readers for text comprehension.
\newblock In \emph{Proceedings of the 55th Annual Meeting of the Association
  for Computational Linguistics (Volume 1: Long Papers)}, pages 1832--1846.
  Association for Computational Linguistics.

\bibitem[{Gururangan et~al.(2018)Gururangan, Swayamdipta, Levy, Schwartz,
  Bowman, and Smith}]{gururangan2018annotation}
Suchin Gururangan, Swabha Swayamdipta, Omer Levy, Roy Schwartz, Samuel Bowman,
  and Noah~A. Smith. 2018.
\newblock Annotation artifacts in natural language inference data.
\newblock In \emph{Proceedings of the 2018 Conference of the North American
  Chapter of the Association for Computational Linguistics: Human Language
  Technologies, Volume 2 (Short Papers)}, pages 107--112. Association for
  Computational Linguistics.

\bibitem[{Hermann et~al.(2015)Hermann, Kocisky, Grefenstette, Espeholt, Kay,
  Suleyman, and Blunsom}]{hermann2015teaching}
Karl~Moritz Hermann, Tomas Kocisky, Edward Grefenstette, Lasse Espeholt, Will
  Kay, Mustafa Suleyman, and Phil Blunsom. 2015.
\newblock Teaching machines to read and comprehend.
\newblock In \emph{Advances in Neural Information Processing Systems (NIPS)},
  pages 1693--1701.

\bibitem[{Hern{\'a}ndez-Orallo(2017)}]{hernandez2017evaluation}
Jos{\'e} Hern{\'a}ndez-Orallo. 2017.
\newblock Evaluation in artificial intelligence: from task-oriented to
  ability-oriented measurement.
\newblock \emph{Artificial Intelligence Review}, 48(3):397--447.

\bibitem[{Hewlett et~al.(2016)Hewlett, Lacoste, Jones, Polosukhin, Fandrianto,
  Han, Kelcey, and Berthelot}]{hewlett2016wikireading}
Daniel Hewlett, Alexandre Lacoste, Llion Jones, Illia Polosukhin, Andrew
  Fandrianto, Jay Han, Matthew Kelcey, and David Berthelot. 2016.
\newblock {WikiReading}: A novel large-scale language understanding task over
  wikipedia.
\newblock In \emph{Proceedings of the 54th Annual Meeting of the Association
  for Computational Linguistics (Volume 1: Long Papers)}, pages 1535--1545,
  Berlin, Germany. Association for Computational Linguistics.

\bibitem[{Hill et~al.(2016)Hill, Bordes, Chopra, and
  Weston}]{hill2015goldilocks}
Felix Hill, Antoine Bordes, Sumit Chopra, and Jason Weston. 2016.
\newblock The goldilocks principle: Reading children's books with explicit
  memory representations.
\newblock In \emph{the International Conference on Learning Representations}.

\bibitem[{Hirschman et~al.(1999)Hirschman, Light, Breck, and
  Burger}]{hirschman1999deep}
Lynette Hirschman, Marc Light, Eric Breck, and John~D. Burger. 1999.
\newblock Deep read: A reading comprehension system.
\newblock In \emph{Proceedings of the 37th Annual Meeting of the Association
  for Computational Linguistics}, pages 325--332. Association for Computational
  Linguistics.

\bibitem[{Jia and Liang(2017)}]{jia2017adversarial}
Robin Jia and Percy Liang. 2017.
\newblock Adversarial examples for evaluating reading comprehension systems.
\newblock In \emph{Proceedings of the 2017 Conference on Empirical Methods in
  Natural Language Processing}, pages 2011--2021, Copenhagen, Denmark.
  Association for Computational Linguistics.

\bibitem[{Joshi et~al.(2017)Joshi, Choi, Weld, and
  Zettlemoyer}]{joshi2017triviaqa}
Mandar Joshi, Eunsol Choi, Daniel Weld, and Luke Zettlemoyer. 2017.
\newblock Trivia{QA}: A large scale distantly supervised challenge dataset for
  reading comprehension.
\newblock In \emph{Proceedings of the 55th Annual Meeting of the Association
  for Computational Linguistics (Volume 1: Long Papers)}, pages 1601--1611.
  Association for Computational Linguistics.

\bibitem[{Khashabi et~al.(2018)Khashabi, Chaturvedi, Roth, Upadhyay, and
  Roth}]{khashabi2018looking}
Daniel Khashabi, Snigdha Chaturvedi, Michael Roth, Shyam Upadhyay, and Dan
  Roth. 2018.
\newblock Looking beyond the surface: A challenge set for reading comprehension
  over multiple sentences.
\newblock In \emph{Proceedings of the 2018 Conference of the North American
  Chapter of the Association for Computational Linguistics: Human Language
  Technologies, Volume 1 (Long Papers)}, pages 252--262. Association for
  Computational Linguistics.

\bibitem[{Ko\v{c}isk\'y et~al.(2018)Ko\v{c}isk\'y, Schwarz, Blunsom, Dyer,
  Hermann, Melis, and Grefenstette}]{kocisky2018narrativeqa}
Tom\'a\v{s} Ko\v{c}isk\'y, Jonathan Schwarz, Phil Blunsom, Chris Dyer,
  Karl~Moritz Hermann, G\'abor Melis, and Edward Grefenstette. 2018.
\newblock The {NarrativeQA} reading comprehension challenge.
\newblock \emph{Transactions of the Association for Computational Linguistics},
  6:317--328.

\bibitem[{Lai et~al.(2017)Lai, Xie, Liu, Yang, and Hovy}]{lai2017race}
Guokun Lai, Qizhe Xie, Hanxiao Liu, Yiming Yang, and Eduard Hovy. 2017.
\newblock {RACE}: Large-scale reading comprehension dataset from examinations.
\newblock In \emph{Proceedings of the 2017 Conference on Empirical Methods in
  Natural Language Processing}, pages 796--805, Copenhagen, Denmark.
  Association for Computational Linguistics.

\bibitem[{Levesque(2014)}]{levesque2014behaviour}
Hector~J. Levesque. 2014.
\newblock On our best behaviour.
\newblock \emph{Artificial Intelligence}, 212:27 -- 35.

\bibitem[{Lin(2004)}]{lin2004rouge}
Chin-Yew Lin. 2004.
\newblock {ROUGE}: A package for automatic evaluation of summaries.
\newblock In \emph{Text Summarization Branches Out: Proceedings of the ACL-04
  Workshop}, pages 74--81, Barcelona, Spain. Association for Computational
  Linguistics.

\bibitem[{LoBue and Yates(2011)}]{lobue2011commonsense}
Peter LoBue and Alexander Yates. 2011.
\newblock Types of common-sense knowledge needed for recognizing textual
  entailment.
\newblock In \emph{Proceedings of the 49th Annual Meeting of the Association
  for Computational Linguistics: Human Language Technologies}, pages 329--334,
  Portland, Oregon, USA. Association for Computational Linguistics.

\bibitem[{Min et~al.(2018)Min, Zhong, Socher, and Xiong}]{min2018efficient}
Sewon Min, Victor Zhong, Richard Socher, and Caiming Xiong. 2018.
\newblock Efficient and robust question answering from minimal context over
  documents.
\newblock In \emph{Proceedings of the 56th Annual Meeting of the Association
  for Computational Linguistics (Volume 1: Long Papers)}, pages 1725--1735.
  Association for Computational Linguistics.

\bibitem[{Mostafazadeh et~al.(2016)Mostafazadeh, Chambers, He, Parikh, Batra,
  Vanderwende, Kohli, and Allen}]{mostafazadeh2016corpus}
Nasrin Mostafazadeh, Nathanael Chambers, Xiaodong He, Devi Parikh, Dhruv Batra,
  Lucy Vanderwende, Pushmeet Kohli, and James Allen. 2016.
\newblock A corpus and cloze evaluation for deeper understanding of commonsense
  stories.
\newblock In \emph{Proceedings of the 2016 Conference of the North American
  Chapter of the Association for Computational Linguistics: Human Language
  Technologies}, pages 839--849, San Diego, California. Association for
  Computational Linguistics.

\bibitem[{Mostafazadeh et~al.(2017)Mostafazadeh, Roth, Louis, Chambers, and
  Allen}]{mostafazadeh2017lsdsem}
Nasrin Mostafazadeh, Michael Roth, Annie Louis, Nathanael Chambers, and James
  Allen. 2017.
\newblock {LSDSem} 2017 shared task: The story cloze test.
\newblock In \emph{Proceedings of the 2nd Workshop on Linking Models of
  Lexical, Sentential and Discourse-level Semantics}, pages 46--51. Association
  for Computational Linguistics.

\bibitem[{Nguyen et~al.(2016)Nguyen, Rosenberg, Song, Gao, Tiwary, Majumder,
  and Deng}]{nguyen2016msmarco}
Tri Nguyen, Mir Rosenberg, Xia Song, Jianfeng Gao, Saurabh Tiwary, Rangan
  Majumder, and Li~Deng. 2016.
\newblock {MS} {MARCO:} {A} human generated machine reading comprehension
  dataset.
\newblock \emph{CoRR}, abs/1611.09268.

\bibitem[{Onishi et~al.(2016)Onishi, Wang, Bansal, Gimpel, and
  McAllester}]{onishi2016who}
Takeshi Onishi, Hai Wang, Mohit Bansal, Kevin Gimpel, and David McAllester.
  2016.
\newblock Who did {W}hat: A large-scale person-centered cloze dataset.
\newblock In \emph{Proceedings of the 2016 Conference on Empirical Methods in
  Natural Language Processing}, pages 2230--2235. Association for Computational
  Linguistics.

\bibitem[{Ostermann et~al.(2018)Ostermann, Modi, Roth, Thater, and
  Pinkal}]{ostermann2018mcscript}
Simon Ostermann, Ashutosh Modi, Michael Roth, Stefan Thater, and Manfred
  Pinkal. 2018.
\newblock {MCS}cript: A novel dataset for assessing machine comprehension using
  script knowledge.
\newblock In \emph{Proceedings of the Eleventh International Conference on
  Language Resources and Evaluation (LREC 2018)}. European Language Resources
  Association (ELRA).

\bibitem[{Paperno et~al.(2016)Paperno, Kruszewski, Lazaridou, Pham, Bernardi,
  Pezzelle, Baroni, Boleda, and Fernandez}]{paperno2016lambada}
Denis Paperno, Germ\'{a}n Kruszewski, Angeliki Lazaridou, Ngoc~Quan Pham,
  Raffaella Bernardi, Sandro Pezzelle, Marco Baroni, Gemma Boleda, and Raquel
  Fernandez. 2016.
\newblock The {LAMBADA} dataset: Word prediction requiring a broad discourse
  context.
\newblock In \emph{Proceedings of the 54th Annual Meeting of the Association
  for Computational Linguistics}, pages 1525--1534. Association for
  Computational Linguistics.

\bibitem[{Rajpurkar et~al.(2018)Rajpurkar, Jia, and Liang}]{rajpurkar2018know}
Pranav Rajpurkar, Robin Jia, and Percy Liang. 2018.
\newblock Know what you don't know: Unanswerable questions for {SQuAD}.
\newblock In \emph{Proceedings of the 56th Annual Meeting of the Association
  for Computational Linguistics (Volume 2: Short Papers)}, pages 784--789,
  Melbourne, Australia. Association for Computational Linguistics.

\bibitem[{Rajpurkar et~al.(2016)Rajpurkar, Zhang, Lopyrev, and
  Liang}]{rajpurkar2016squad}
Pranav Rajpurkar, Jian Zhang, Konstantin Lopyrev, and Percy Liang. 2016.
\newblock {SQ}u{AD}: 100,000+ questions for machine comprehension of text.
\newblock In \emph{Proceedings of the 2016 Conference on Empirical Methods in
  Natural Language Processing}, pages 2383--2392. Association for Computational
  Linguistics.

\bibitem[{Richardson et~al.(2013)Richardson, Burges, and
  Renshaw}]{richardson2013MCTest}
Matthew Richardson, Christopher~J.C. Burges, and Erin Renshaw. 2013.
\newblock {MCT}est: A challenge dataset for the open-domain machine
  comprehension of text.
\newblock In \emph{Proceedings of the 2013 Conference on Empirical Methods in
  Natural Language Processing}, pages 193--203.

\bibitem[{Sachan et~al.(2015)Sachan, Dubey, Xing, and
  Richardson}]{sachan2015learning}
Mrinmaya Sachan, Kumar Dubey, Eric Xing, and Matthew Richardson. 2015.
\newblock Learning answer-entailing structures for machine comprehension.
\newblock In \emph{Proceedings of the 53rd Annual Meeting of the Association
  for Computational Linguistics and the 7th International Joint Conference on
  Natural Language Processing (Volume 1: Long Papers)}, pages 239--249.
  Association for Computational Linguistics.

\bibitem[{Saha et~al.(2018)Saha, Aralikatte, Khapra, and
  Sankaranarayanan}]{saha2018duorc}
Amrita Saha, Rahul Aralikatte, Mitesh~M. Khapra, and Karthik Sankaranarayanan.
  2018.
\newblock Duo{RC}: Towards complex language understanding with paraphrased
  reading comprehension.
\newblock In \emph{Proceedings of the 56th Annual Meeting of the Association
  for Computational Linguistics (Volume 1: Long Papers)}, pages 1683--1693,
  Melbourne, Australia. Association for Computational Linguistics.

\bibitem[{Sammons et~al.(2010)Sammons, Vydiswaran, and Roth}]{sammons2010ask}
Mark Sammons, V.G.Vinod Vydiswaran, and Dan Roth. 2010.
\newblock ``{A}sk not what textual entailment can do for you...''.
\newblock In \emph{Proceedings of the 48th Annual Meeting of the Association
  for Computational Linguistics}, pages 1199--1208. Association for
  Computational Linguistics.

\bibitem[{Schwartz et~al.(2017)Schwartz, Sap, Konstas, Zilles, Choi, and
  Smith}]{schwartz2017story}
Roy Schwartz, Maarten Sap, Ioannis Konstas, Leila Zilles, Yejin Choi, and
  Noah~A. Smith. 2017.
\newblock Story cloze task: {UW NLP} system.
\newblock In \emph{Proceedings of the 2nd Workshop on Linking Models of
  Lexical, Sentential and Discourse-level Semantics}, pages 52--55, Valencia,
  Spain. Association for Computational Linguistics.

\bibitem[{Seo et~al.(2017)Seo, Kembhavi, Farhadi, and
  Hajishirzi}]{seo2017bidirectional}
Minjoon Seo, Aniruddha Kembhavi, Ali Farhadi, and Hannaneh Hajishirzi. 2017.
\newblock Bidirectional attention flow for machine comprehension.
\newblock In \emph{International Conference on Learning Representations}.

\bibitem[{Sugawara et~al.(2017)Sugawara, Kido, Yokono, and
  Aizawa}]{sugawara2017evaluation}
Saku Sugawara, Yusuke Kido, Hikaru Yokono, and Akiko Aizawa. 2017.
\newblock Evaluation metrics for machine reading comprehension: Prerequisite
  skills and readability.
\newblock In \emph{Proceedings of the 55th Annual Meeting of the Association
  for Computational Linguistics (Volume 1: Long Papers)}, pages 806--817.
  Association for Computational Linguistics.

\bibitem[{Suster and Daelemans(2018)}]{suster2018clicr}
Simon Suster and Walter Daelemans. 2018.
\newblock Cli{CR}: a dataset of clinical case reports for machine reading
  comprehension.
\newblock In \emph{Proceedings of the 2018 Conference of the North American
  Chapter of the Association for Computational Linguistics: Human Language
  Technologies, Volume 1 (Long Papers)}, pages 1551--1563. Association for
  Computational Linguistics.

\bibitem[{Sutcliffe et~al.(2013)Sutcliffe, Pe{\~n}as, Hovy, Forner, Rodrigo,
  Forascu, Benajiba, and Osenova}]{sutcliffe2013QA4MRE}
Richard Sutcliffe, Anselmo Pe{\~n}as, Eduard Hovy, Pamela Forner, {\'A}lvaro
  Rodrigo, Corina Forascu, Yassine Benajiba, and Petya Osenova. 2013.
\newblock Overview of {QA4MRE} main task at {CLEF} 2013.
\newblock \emph{Working Notes, CLEF}.

\bibitem[{Trischler et~al.(2017)Trischler, Wang, Yuan, Harris, Sordoni,
  Bachman, and Suleman}]{trischler2017newsqa}
Adam Trischler, Tong Wang, Xingdi Yuan, Justin Harris, Alessandro Sordoni,
  Philip Bachman, and Kaheer Suleman. 2017.
\newblock News{QA}: A machine comprehension dataset.
\newblock In \emph{Proceedings of the 2nd Workshop on Representation Learning
  for NLP}, pages 191--200. Association for Computational Linguistics.

\bibitem[{Weissenborn et~al.(2017)Weissenborn, Wiese, and
  Seiffe}]{weissenborn2017making}
Dirk Weissenborn, Georg Wiese, and Laura Seiffe. 2017.
\newblock Making neural {QA} as simple as possible but not simpler.
\newblock In \emph{Proceedings of the 21st Conference on Computational Natural
  Language Learning (CoNLL 2017)}, pages 271--280, Vancouver, Canada.
  Association for Computational Linguistics.

\bibitem[{Welbl et~al.(2018)Welbl, Stenetorp, and
  Riedel}]{welbl2018constructing}
Johannes Welbl, Pontus Stenetorp, and Sebastian Riedel. 2018.
\newblock Constructing datasets for multi-hop reading comprehension across
  documents.
\newblock \emph{Transactions of the Association for Computational Linguistics},
  6:287--302.

\bibitem[{Weston et~al.(2015)Weston, Bordes, Chopra, and
  Mikolov}]{weston2015bAbI}
Jason Weston, Antoine Bordes, Sumit Chopra, and Tomas Mikolov. 2015.
\newblock Towards {AI}-complete question answering: a set of prerequisite toy
  tasks.
\newblock In \emph{the International Conference on Learning Representations}.

\bibitem[{Whiteson et~al.(2011)Whiteson, Tanner, Taylor, and
  Stone}]{whiteson2011protecting}
Shimon Whiteson, Brian Tanner, Matthew~E Taylor, and Peter Stone. 2011.
\newblock Protecting against evaluation overfitting in empirical reinforcement
  learning.
\newblock In \emph{Adaptive Dynamic Programming And Reinforcement Learning
  (ADPRL), 2011 IEEE Symposium on}, pages 120--127. IEEE.

\bibitem[{Yin et~al.(2016)Yin, Ebert, and Sch\"{u}tze}]{yin2016HABCNN}
Wenpeng Yin, Sebastian Ebert, and Hinrich Sch\"{u}tze. 2016.
\newblock Attention-based convolutional neural network for machine
  comprehension.
\newblock In \emph{Proceedings of the Workshop on Human-Computer Question
  Answering}, pages 15--21. Association for Computational Linguistics.

\bibitem[{Yu et~al.(2018)Yu, Dohan, Le, Luong, Zhao, and Chen}]{yu2018qanet}
Adams~Wei Yu, David Dohan, Quoc Le, Thang Luong, Rui Zhao, and Kai Chen. 2018.
\newblock {QAN}et: Combining local convolution with global self-attention for
  reading comprehension.
\newblock In \emph{International Conference on Learning Representations}.

\end{thebibliography}
\bibliographystyle{acl_natbib_nourl}

\appendix

\section{Hyperparameters of the Baseline Systems}
\label{sec:app1}

We used different hyperparameters for each dataset owing to the difference characteristics of the datasets, e.g., the context length. Tables \ref{tbl:hyper1} and \ref{tbl:hyper2} show the hyperparameters.

\begin{table}[th]
  \centering
  \newcommand{\linestack}[1]{    \def\arraystretch{0.9}\begin{tabular}[c]{@{}c@{}} #1 \end{tabular}}
  \begin{tabular}{lccccccc} \toprule
    Dataset & $b$ & $h$  & $q$ & $d$ \\ \midrule
    SQuAD    & 60 & 100 & 400  & 20  \\
    AddSent  & 60 & 100 & 400  & 20  \\
    NewsQA   & 32 & 100 & 1000 & 20  \\
    TriviaQA & 32 & 100 & 400  & 20  \\
    QAngaroo & 16 & 50  & 4096 & 20  \\
    MARCO    & 20 & 40  & 1600 & 30  \\
  NarrativeQA& 60 & 50  & 1000 & 20  \\ \bottomrule
  \end{tabular}
  \caption{Hyperparameters (batch size $b$, hidden layer size $h$, document size threshold $d$, question size threshold $q$) of the Bidirectional Attention Flow \cite{seo2017bidirectional} for each dataset. The other settings basically follow the original implementation. In TriviaQA, we followed a method for the dataset preparation used in \newcite{joshi2017triviaqa}.}
  \label{tbl:hyper1}
\end{table}

\begin{table}[h]
  \centering
  \newcommand{\linestack}[1]{    \def\arraystretch{0.9}\begin{tabular}[c]{@{}c@{}} #1 \end{tabular}}
  \begin{tabular}{lccccccc} \toprule
    Dataset & $b$ & $h$ & $n$ & $dr$ & $lr$ \\ \midrule
  MCTest   & 10 & 32  & 1 & 0.5 & 0.01 \\
  RACE     & 32 & 128 & 1 & 0.2 & 0.1  \\
  MCScript & 25 & 64  & 1 & 0.5 & 0.2  \\
  ARC-E    & 32 & 256 & 1 & 0.5 & 0.3  \\
  ARC-C    & 32 & 256 & 1 & 0.5 & 0.3  \\ \bottomrule
  \end{tabular}
  \caption{Hyperparameters (batch size $b$, hidden layer size $h$, number of attention layers $n$, dropout rate $dr$, learning rate $lr$) of the Gated-Attentive Reader \cite{dhingra2017gated} for each dataset. The other settings basically follow as the implementation in \newcite{lai2017race}.}
  \label{tbl:hyper2}
\end{table}

\end{document}